\documentclass[letterpaper, 10 pt, journal, twoside]{IEEEtran}

\usepackage{graphics} 
\usepackage{epsfig} 
\usepackage{epstopdf}
\graphicspath{ {figs} }
\usepackage{caption}
\usepackage{xcolor}
\usepackage{subcaption}
\usepackage{lipsum}
\usepackage{float}
\usepackage{todonotes}
\usepackage{color,soul}
\usepackage{hyperref}
\usepackage{multirow}
\usepackage{algorithm}
\usepackage{algorithmic}

\usepackage{booktabs}
\usepackage{times}
\usepackage{xcolor}
\usepackage{soul}
\usepackage[utf8]{inputenc}
\usepackage{mwe,tikz}
\usepackage[percent]{overpic}
\usepackage{xcolor}
\usepackage[export]{adjustbox}
\usepackage{tikz}
\usepackage{hyperref}
\usepackage{graphicx}
\usepackage{graphics}
\usepackage{amsmath}
\usepackage{amssymb}
\usepackage{subcaption}
\usepackage{soul}
\usepackage{multirow}
\usepackage{array}
\usepackage{bbm}
\usepackage[most]{tcolorbox}
\usepackage{nth}
\usepackage{hepunits}
\usepackage{stfloats}
\usepackage{cite}

\usepackage{amsmath} 

\usepackage{amssymb}  
\usepackage{array}
\newcolumntype{x}[1]{>{\centering\arraybackslash\hspace{0pt}}p{#1}}

\newcommand{\bY}{\mathbf{Y}}

\definecolor{dblue}{rgb}{0,0,0.7}

\begin{document}

\title{VR-Goggles for Robots: \\Real-to-sim Domain Adaptation for Visual Control}

\author{
  Jingwei Zhang$^{*1}$ \
  Lei Tai$^{*2}$ \
  Peng Yun$^{2}$ \
  Yufeng Xiong$^{1}$ \
  Ming Liu$^{2}$ \
  Joschka Boedecker$^{1}$ \
  Wolfram Burgard$^{1}$
  \thanks{This work was supported by:
  the Shenzhen Science, Technology and Innovation Commission (SZSTI) JCYJ20160428154842603;
  the BrainLinks-BrainTools cluster of excellence funded by the DFG (German Research Foundation), grant number EXC 1086;
  the Research Grant Council of Hong Kong SAR Government, China, under Project No. 11210017, No. 16212815 and No. 21202816, the National Natural Science Foundation of China (Grant No. U1713211) awarded to Prof. Ming Liu.
  \textit{(Corresponding authors: Jingwei Zhang and Lei Tai.)}}
  \thanks{$^{*}$The first two authors contributed equally to this work.}
  \thanks{$^{1}$Jingwei Zhang, Yufeng Xiong, Joschka Boedecker and Wolfram Burgard are with
  the Department of Computer Science,
  University of Freiburg. Breisgau 79110, Germany
  (e-mail: {\tt\footnotesize \{zhang, xiongy, jboedeck, burgard\}@informatik.uni-freiburg.de})}
  \thanks{$^{2}$Lei Tai, Peng Yun and Ming Liu are with
  the Department of Electronic and Computer Engineering,
  The Hong Kong University of Science and Technology, Hong Kong
  (e-mail: {\tt\footnotesize \{ltai, pyun, eelium\}@ust.hk})}
}


\maketitle

\begin{abstract}
  In this paper, we deal with the \textit{reality gap} from a novel
  perspective, targeting transferring Deep Reinforcement Learning
  (DRL) policies learned in simulated environments to the real-world
  domain for visual control tasks. Instead of adopting the common
  solutions to the problem by increasing the visual fidelity of
  synthetic images output from simulators during the training phase,
  we seek to tackle the problem by translating the real-world image
  streams back to the synthetic domain during the deployment phase, to
  \textit{make the robot feel at home}. We propose this as a
  lightweight, flexible, and efficient solution for visual control, as
  1) no extra transfer steps are required during the expensive
  training of DRL agents in simulation; 2) the trained DRL agents will
  not be constrained to being deployable in only one specific
  real-world environment; 3) the policy training and the transfer
  operations are decoupled, and can be conducted in parallel. Besides
  this, we propose a simple yet effective \textit{shift loss} that is agnostic to the downstream task, to
  constrain the consistency between subsequent frames which is
  important for consistent policy outputs.  We validate the
  \textit{shift loss} for \textit{artistic style transfer for videos}
  and \textit{domain adaptation}, and validate our visual control
  approach in indoor and outdoor robotics experiments.
\end{abstract}


\IEEEpeerreviewmaketitle

\section{Introduction}
\label{sec:introduction}

\IEEEPARstart{P}{ioneered} by the Deep Q-network \cite{mnih2015human} and followed up by various extensions and advancements
\cite{mnih2016asynchronous,lillicrap2015continuous,schulman2015trust,schulman2017proximal},
Deep Reinforcement Learning (DRL) algorithms show great potential in solving high-dimensional real-world robotics sensory control tasks.
However, DRL methods typically require several millions of training samples, making them infeasible to train directly on real robotic systems.
As a result, DRL algorithms are generally trained in simulated environments, then transferred to and deployed in real scenes.
However, the \textit{reality gap},
namely the noise pattern, texture, lighting condition discrepancies, etc.,
between synthetic renderings and real sensory readings, imposes major challenges for generalising the sensory control policies trained in simulation to reality.

In this paper, we focus on visual control tasks, where autonomous agents perceive the environment with their onboard cameras, and execute commands based on the colour image reading streams.
A natural way and also the typical choice in the recent literature on dealing with the \textit{reality gap} for visual control,
is by increasing the visual fidelity of the simulated images \cite{bousmalis2017using,stein2018genesis}, by matching the distribution of synthetic images to that of the real ones \cite{Sadeghi2017cadrl,tobin2017domain},
and by gradually adapting the learned features and representations from the simulated domain to the real-world domain \cite{rusu2016sim}.
These \textit{sim-to-real} methods, however, inevitably have to add preprocessing steps for each individual training frame to the already expensive learning pipeline of DRL policies;
or a policy training or finetuning phase has to be conducted for each visually different real-world scene.

This paper attempts to tackle the \textit{reality gap} in the visual control domain from a novel perspective,
with the aim of adding minimal extra computational burden to the learning pipeline.
We cope with the \textit{reality gap} only during the actual deployment phase of agents in real-world scenarios, by adapting the real camera streams to the synthetic modality,
so as to translate the unfamiliar or unseen features of real images back into the simulated style, which the agents have already learned how to deal with during training in the simulation.

Compared to the \textit{sim-to-real} methods bridging the \textit{reality gap}, our proposed \textit{real-to-sim} approach, which we refer to as the \textit{VR-Goggles}, has several appealing properties:
  (1) Our proposed method is highly lightweight:
  It does not add any extra processing burden to the training phase of DRL policies;
  and (2) Our approach is highly flexible and efficient:
  Since we decouple the policy training and the adaptation operations, the preparations for transferring the polices from simulation to the real world can be conducted in parallel with the training of the control policies.
  From each visually different real-world environment that we expect to deploy the agent in, we just need to collect several (typically on the order of $2000$) images, and train a \textit{VR-Goggles} model for each of them.
  More importantly, we do not need to retrain or finetune the visual control policy for new environments.

As an additional contribution,
  we propose a new \textit{shift loss}, which enables generating consistent synthetic image streams without imposing temporal constraints, and does not require sequential training data.
  We show that \textit{shift loss} is a promising and cheap alternative to the constraints imposed by optical flow, and demonstrate its effectiveness in \textit{artistic style transfer for videos} and \textit{domain adaptation}.

\section{Related Works}
\label{sec:relatedWorks}

\subsection{Domain Adaptation}
\label{sec:related-da}
Visual \textit{domain adaptation}, or \textit{image-to-image translation}, targets translating images from a source domain into a target domain.
We here focus on the most general unsupervised methods that require minimal manual effort and are applicable in robotics control tasks.

\textit{CycleGAN}~\cite{zhu2017unpaired} introduced a cycle-consistent loss to enforce an inverse mapping from the target domain to the source domain on top of the source to target mapping.  It does not require paired data from the two domains of interest and shows convincing results for relatively simple data distributions containing few semantic types.  However, in terms of translating between more complex data distributions containing many more semantic types,
its results are not as satisfactory, in that permutations of semantics often occur.
Several works investigate imposing semantic constraints \cite{hoffman2017cycada, cherian2018sem}, e.g., \textit{CyCADA} \cite{hoffman2017cycada} enforces a matching between the semantic map of the translated image and that of the input.

\subsection{\textit{Domain Adaptation} for Learning based Visual Control}
\label{sec:related-da-drl}
Learning-based methods
such as DRL and imitation learning have been applied to robotics control tasks including manipulation and navigation.
Below we review the recent literature mainly
considering the visual \textit{reality gap}.

Bousmalis et al. \cite{bousmalis2017using} bridged the \textit{reality gap} for manipulation by adapting synthetic images to the realistic domain during training, with a combination of image-level and feature-level adaptation.
Also following the \textit{sim-to-real} direction, Stein et al. \cite{stein2018genesis} utilized \textit{CycleGAN} to translate every synthetic frame to the realistic style during training navigation policies.
Although effective, these approaches still add an adaptation step before each training iteration, which can slow down the whole learning pipeline.

The method of domain randomization \cite{Sadeghi2017cadrl,tobin2017domain,pinto2017asymmetric} is proposed to randomize the texture of objects, lighting conditions, and camera positions during training, such that the learned model could generalize naturally to real-world scenarios. However, such randomizing might not be efficiently realized by some robotic simulators at a relatively low cost. Moreover, there is no guarantee that these randomized simulations can cover the visual modality of an arbitrary real-world scene.

Rusu et al. \cite{rusu2016sim} deals with the \textit{reality gap} by progressively adapting the features and representations learned in simulation to that of the realistic domain. This method, however, still needs to go through a policy finetuning phase for each visually different real-world scenario.

Apart from the approaches mentioned above, some works chose special setups to circumvent the \textit{reality gap}. For example, $2D$ Lidar~\cite{tai2017virtual,zhang2017neural,zhelo2018curiosity} and depth images~\cite{zhang2016deep,tai2018social} are sometimes chosen as the sensor modality,
since the discrepancies between the simulated domain and the real-world domain for them can be smaller than those for colour images. Zhu et al. \cite{zhu2017target} conducted real-world experiments with visual inputs. However, in their setups, the real-world scene is highly visually similar to the simulation, a condition that can be relatively difficult to meet in practice.

Very related to our method is the work of Inoue et al. which also adopts a \textit{real-to-sim} direction \cite{inoue2018transfer}. They train VAEs to perform the adaptation during deployment of the trained object detection model in the real world. However, their method relies on paired data between two domains and focuses on supervised perception tasks.

In this paper, we mainly consider \textit{domain adaptation} for learning-based visual navigation. In terms of visual aspects, the adaptation for navigation is quite challenging, since navigation agents usually work in environments at relatively larger scales compared to the relatively confined workspaces for manipulators. We believe our proposed \textit{real-to-sim} method could be potentially adopted in other control domains.

An essential aspect of \textit{domain adaptation}, within the context of dealing with the \textit{reality gap} is the consistency between subsequent frames, which has not been considered in any of the adaptation methods mentioned above. As an approach for solving sequential decision making, the consistency between the subsequent inputs for DRL agents can be critical for the successful fulfilment of their final goals. Apart from solutions for solving the \textit{reality gap}, the general \textit{domain adaptation} literature also lacks works considering sequential frames instead of single frames. Therefore, we look to borrow techniques from other fields that successfully extend single-frame algorithms to the video domain, among which the most applicable methods are from the \textit{artistic style transfer} literature.

\subsection{Artistic Style Transfer for Videos}
\label{sec:related-style}

\textit{Artistic style transfer} is a technique for transferring the artistic style of artworks to photographs \cite{johnson2016perceptual}.
\textit{Artistic style transfer for videos} works on video sequences instead of individual frames, targeting generating temporally consistent stylizations for sequential inputs.
Ruder et al. \cite{ruder2017artistic} provides a key observation that: a trained stylization network with a total downsampling factor of $K$ (e.g., $K=4$ for a network with $2$ convolutional layers of stride $2$),
is shift invariant to shifts equal to the multiples of $K$ pixels, but can output substantially different stylizations otherwise.
This undesired property (of not being shift invariant) causes the output of the trained network to change substantially for even very tiny changes in the input, which leads to temporal inconsistency (under the assumption that only relatively limited changes would appear in subsequent input frames).
However, their solution of adding temporal constraints between generated subsequent frames, is rather expensive, as it requires optical flow as input during deployment.
Huang et al. \cite{huang2017real} offers a relatively cheap solution, requiring the temporal constraint only during training single-frame \textit{artistic style transfer}.
However, we suspect that constraining optical flow on single frames is not well-defined.
We suspect that their improved temporal consistency
is actually due to the inexplicitly imposed consistency constraints for regional shifts by optical flow.
We validate this suspicion in our experiments (Sec. \ref{sec:experiments-style}).

\begin{figure*}[!ht]
    \centering
    \includegraphics[width=0.82\textwidth]{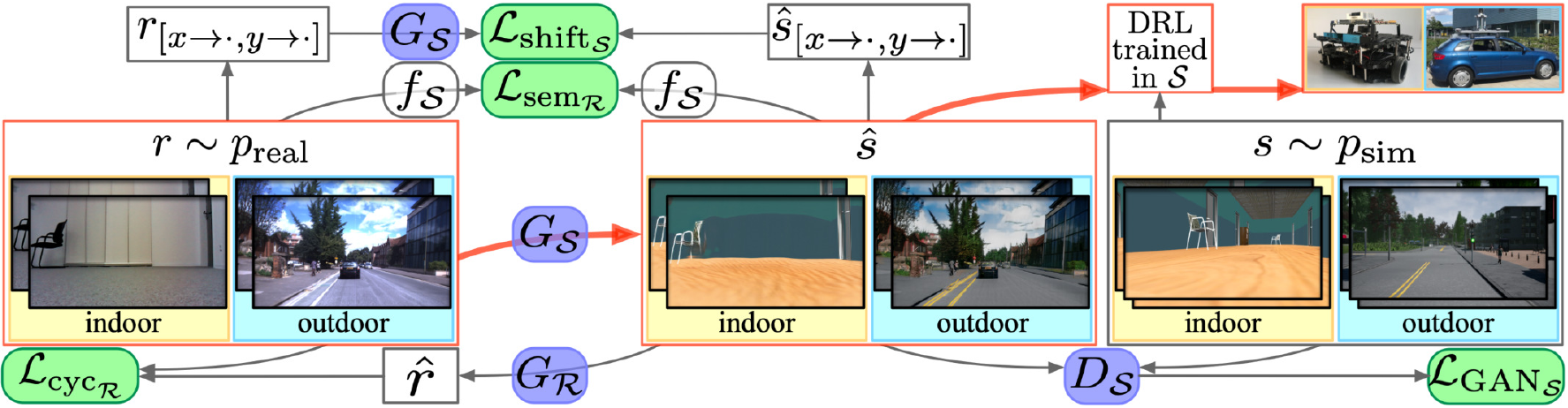}
    \caption{The \textit{VR-Goggles} pipeline. We depict the computation of the losses $\mathcal{L}_{\text{GAN}_{\mathcal{S}}}$,
    $\mathcal{L}_{\text{cyc}_{\mathcal{R}}}$, $\mathcal{L}_{\text{sem}_{\mathcal{R}}}$ and $\mathcal{L}_{\text{shift}_{\mathcal{S}}}$.
    We present both \textit{outdoor} and \textit{indoor} scenarios,
    where the adaptation for the \textit{outdoor} scene is trained with the semantic loss $\mathcal{L}_{\text{sem}}$ (since its simulated domain \textit{CARLA}
    has ground truth semantic labels to train a segmentation network $f_{\mathcal{S}}$),
    and the \textit{indoor} one without (since its simulated domain \textit{Gazebo}
    does not provide semantic ground truth).
    The components marked in \textit{red} are those involved in the final deployment: a real sensor reading is captured ($r\sim p_{\text{real}}$),
    then passed through the generator $G_{\mathcal{S}}$ to be translated into the simulated domain $\mathcal{S}$, where the DRL agents were originally trained;
    the translated image $\hat{s}$ is then fed to the DRL policy, which outputs control commands.
    For clarity, we skip the counterpart losses $\mathcal{L}_{\text{GAN}_{\mathcal{R}}}$, $\mathcal{L}_{\text{cyc}_{\mathcal{S}}}$, $\mathcal{L}_{\text{sem}_{\mathcal{S}}}$ and $\mathcal{L}_{\text{shift}_{\mathcal{R}}}$.
    }
    \label{fig:vr-goggles}
    \vspace{-0.02in}
\end{figure*}

We propose that the fundamental problem causing the inconsistency
can be solved by an additional constraint of \textit{shift loss},
which we introduce in Sec. \ref{sec:shift-loss}.
We show that the \textit{shift loss} constrains the consistency between generated subsequent frames,
without the need for the relatively expensive optical flow constraint.
We argue that for a network that has been properly trained to learn a smooth function approximation, small changes in the input should also result in small changes in the output.

\section{Methods}
\label{sec:methods}

\subsection{Problem formulation}
We consider visual data sources from two domains: $\mathcal{S}$, containing sequential frames $\{s_0,s_1,s_2,\cdots\}$ (e.g., synthetic images output from a simulator; $s\sim p_{\text{sim}}$, where $p_{\text{sim}}$ denotes the simulated data distribution), and $\mathcal{R}$, containing sequential frames $\{r_0,r_1,r_2,\cdots\}$ (e.g., real camera readings from the onboard camera of a mobile robot; $r\sim p_{\text{real}}$, where $p_{\text{real}}$ denotes the distribution of the real sensory readings). We emphasize that, although we require our method to generate consistent outputs for sequential inputs, we do not need the training data to be sequential; we formalize it in this way only because some of our baseline methods have this requirement.

DRL agents are typically trained in the simulated domain $\mathcal{S}$,
and expected to execute in the real-world domain $\mathcal{R}$.
As we have discussed, we choose to tackle this problem by translating the images from $\mathcal{R}$
to $\mathcal{S}$
during deployment.
In the following, we introduce our approach for performing \textit{domain adaptation}.
Also to cope with the sequential nature of the incoming data streams, we introduce a \textit{shift loss} technique for constraining the consistency of the translated subsequent frames.

\subsection{\textit{CycleGAN} Loss}

We first build on top of \textit{CycleGAN} \cite{zhu2017unpaired}, which learns two generative models to map between domains:
$G_{\mathcal{R}}: \mathcal{S}\rightarrow \mathcal{R}$, with its discriminator $D_{\mathcal{R}}$, and $G_{\mathcal{S}}: \mathcal{R}\rightarrow \mathcal{S}$, with its discriminator $D_{\mathcal{S}}$, via training two GANs simultaneously:
\begin{align}
\mathcal{L}_{\text{GAN}_{\mathcal{R}}}(G_{\mathcal{R}},D_{\mathcal{R}};\mathcal{S},\mathcal{R})
=& \mathbb{E}_{p_{\text{real}}} \left[ \log D_{\mathcal{R}}(r) \right] +
\nonumber\\
& \mathbb{E}_{p_{\text{sim}}} \left[ \log (1-D_{\mathcal{R}}(G_{\mathcal{R}}(s))) \right],
\nonumber\\
\mathcal{L}_{\text{GAN}_{\mathcal{S}}}(G_{\mathcal{S}},D_{\mathcal{S}};\mathcal{R},\mathcal{S})
=&\mathbb{E}_{p_{\text{sim}}} \left[ \log D_{\mathcal{S}}(s) \right] +
\nonumber\\
& \mathbb{E}_{p_{\text{real}}} \left[ \log (1-D_{\mathcal{S}}(G_{\mathcal{S}}(r))) \right], \nonumber
\end{align}
in which $G_{\mathcal{R}}$ learns to generate images $G_{\mathcal{R}}(s)$ matching those from domain $\mathcal{R}$, while $G_{\mathcal{S}}$ translats $r$ to domain $\mathcal{S}$.
We also constrain mappings with the \textit{cycle consistency loss} \cite{zhu2017unpaired}:
\begin{align}
    \mathcal{L}_{\text{cyc}_{\mathcal{R}}}(G_{\mathcal{S}},G_{\mathcal{R}};\mathcal{R})
&=
    \mathbb{E}_{p_{\text{real}}}\left[ \left|\left| G_{\mathcal{R}}(G_{\mathcal{S}}(r))-r \right|\right|_{1} \right],
\nonumber\\
    \mathcal{L}_{\text{cyc}_{\mathcal{S}}}(G_{\mathcal{R}},G_{\mathcal{S}};\mathcal{S})
&=
    \mathbb{E}_{p_{\text{sim}}}\left[ \left|\left| G_{\mathcal{S}}(G_{\mathcal{R}}(s))-s \right|\right|_{1} \right].
\nonumber
\end{align}

\subsection{Semantic Loss}

Since our translation domains of interest are between synthetic images and real-world sensor images, we take advantage of the fact that many recent robotic simulators provide ground truth semantic labels and add a semantic constraint inspired by \textit{CyCADA} \cite{hoffman2017cycada}.
(For simplicity in the following we use \textit{CyCADA} to refer to \textit{CycleGAN} plus this semantic loss instead of the full \textit{CyCADA} approach \cite{hoffman2017cycada}).

Assuming that for images from domain $\mathcal{S}$, the ground truth semantic labels $\bY$ are available, a semantic segmentation network $f_{\mathcal{S}}$ can be obtained by minimizing the \textit{cross-entropy} loss $\mathbb{E}_{s\sim\mathcal{S}}[\text{CrossEnt}(\bY_{s},f_{\mathcal{S}}(s))]$.
We further assume that the ground truth semantic for domain $\mathcal{R}$ is lacking (which is the case for most real scenarios), meaning that $f_{\mathcal{R}}$ is not easily obtainable.
In this case, we use $f_{\mathcal{S}}$ to generate "semi" semantic labels for domain $\mathcal{R}$.
Then semantically consistent image translation can be achieved by minimizing the following losses,
which imposes consistency between the semantic maps of the input and that of the generated output:
\begin{align}
    \mathcal{L}_{\text{sem}_{\mathcal{R}}}(G_{\mathcal{S}};\mathcal{R},f_{\mathcal{S}})&
=
    \mathbb{E}_{p_{\text{real}}}[\text{CrossEnt}(f_{\mathcal{S}}(r),f_{\mathcal{S}}(G_{\mathcal{S}}(r)))].
\nonumber\\
    \mathcal{L}_{\text{sem}_{\mathcal{S}}}(G_{\mathcal{R}};\mathcal{S},f_{\mathcal{S}})&
=
    \mathbb{E}_{p_{\text{sim}}}[\text{CrossEnt}(f_{\mathcal{S}}(s),f_{\mathcal{S}}(G_{\mathcal{R}}(s)))],
\nonumber
\end{align}

\subsection{\textit{Shift Loss} for Consistent Generation}
\label{sec:shift-loss}

Different from the current literature of
\textit{domain adaptation},
our model is additionally expected to output consistent images for sequential inputs.
Although with $\mathcal{L}_{\text{sem}}$, the semantics of the consecutive outputs are constrained, inconsistencies and artifacts still occur quite often.
Moreover, in cases where ground truth semantics are unavailable from either domain,
the sequential outputs are even less constrained,
which could potentially lead to inconsistent policy outputs.
Following the discussions in Sec. \ref{sec:related-style},
we introduce the \textit{shift loss}
to constrain the consistency even in these situations.

For an input image $s$, we use $s_{\left[x\rightarrow i, y\rightarrow j\right]}$ to denote the result of a shift operation: shifting $s$ along the $X$ axis by $i$ pixels, and $j$ pixels along the $Y$ axis.
We sometimes omit $y\rightarrow0$ or $x\rightarrow0$ in the subscript if the image is only shifted along the $X$ or $Y$ axis.
According to \cite{ruder2017artistic}, a trained stylization network is shift invariant to shifts of multiples of $K$ pixels ($K$ represents the total downsampling factor of the network), but can output significantly different stylizations otherwise. This causes the output of the trained network to change greatly for even very small changes in the input.
We thus propose to add a simple yet direct and effective \textit{shift loss}
($u$ denotes uniform distribution):
\begin{align}
  \mathcal{L}_{\text{shift}_{\mathcal{R}}}(G_{\mathcal{R}}; \mathcal{S}) &=
  \mathbb{E}_{p_{\text{sim}}, \ i,j\thicksim u(1, K-1) } \nonumber \\
  & \left[\left|\left|G_{\mathcal{R}}(s)_{\left[x\rightarrow i, y\rightarrow j\right]} - G_{\mathcal{R}}(s_{\left[x\rightarrow i, y\rightarrow j\right]})\right|\right|_{2}^{2}\right],
  \nonumber\\
  \mathcal{L}_{\text{shift}_{\mathcal{S}}}(G_{\mathcal{S}}; \mathcal{R}) &=
  \mathbb{E}_{p_{\text{real}}, \ i,j\thicksim u(1, K-1)} \nonumber \\
  & \left[\left|\left|G_{\mathcal{S}}(r)_{\left[x\rightarrow i, y\rightarrow j\right]} - G_{\mathcal{S}}(r_{\left[x\rightarrow i, y\rightarrow j\right]})\right|\right|_{2}^{2}\right].
  \nonumber
\end{align}

\textit{Shift loss} constrains the shifted output to match the output of the shifted input,
regarding the shifts as image-scale movements.
Assuming that only limited regional movement would appear in subsequent input frames,
\textit{shift loss} effectively smoothes the mapping function for small regional movements,
restricting the changes in its outputs for subsequent inputs.
This can be regarded as a cheap alternative for imposing consistency constraints on small movements,
eliminating the need for the optical flow information,
which is crucial for meeting the requirements of real-time robotics control.

\subsection{Full Objective}

Our full objective for learning \textit{VR-Goggles} (Fig. \ref{fig:vr-goggles}) is
($\lambda_{\text{cyc}}$, $\lambda_{\text{sem}}$ and $\lambda_{\text{shift}}$ are the loss weightings):
\begin{align}
\mathcal{L}(& G_{\mathcal{R}}, G_{\mathcal{S}},D_{\mathcal{R}},D_{\mathcal{S}};\mathcal{S},\mathcal{R},f_{\mathcal{S}})
\nonumber
\\&=
    \mathcal{L}_{\text{GAN}_{\mathcal{R}}}(G_{\mathcal{R}},D_{\mathcal{R}};\mathcal{S},\mathcal{R})
+
    \mathcal{L}_{\text{GAN}_{\mathcal{S}}}(G_{\mathcal{S}},D_{\mathcal{S}};\mathcal{R},\mathcal{S})
\nonumber
\\\nonumber&+
    \lambda_{\text{cyc}}
    \left(
    \mathcal{L}_{\text{cyc}_{\mathcal{R}}}(G_{\mathcal{S}},G_{\mathcal{R}};\mathcal{R})
+
    \mathcal{L}_{\text{cyc}_{\mathcal{S}}}(G_{\mathcal{R}},G_{\mathcal{S}};\mathcal{S})
    \right)
\\\nonumber&+
    \lambda_{\text{sem}}
    \left(
    \mathcal{L}_{\text{sem}_{\mathcal{R}}}(G_{\mathcal{S}};\mathcal{R},f_{\mathcal{R}})
+
    \mathcal{L}_{\text{sem}_{\mathcal{S}}}(G_{\mathcal{R}};\mathcal{S},f_{\mathcal{S}})
    \right)
\\\nonumber&+
    \lambda_{\text{shift}}
    \left(
    \mathcal{L}_{\text{shift}_{\mathcal{R}}}(G_{\mathcal{R}};\mathcal{S})
+
    \mathcal{L}_{\text{shift}_{\mathcal{S}}}(G_{\mathcal{S}};\mathcal{R})
    \right).
\end{align}
This corresponds to solving the following optimization:
\begin{align}
    G_{\mathcal{R}}^*, G_{\mathcal{S}}^*
&=
    \arg\min_{G_{\mathcal{R}},G_{\mathcal{S}}}\max_{D_{\mathcal{R}},D_{\mathcal{S}}}\mathcal{L}(G_{\mathcal{R}},G_{\mathcal{S}},D_{\mathcal{R}},D_{\mathcal{S}}). \nonumber
\end{align}

\section{Experiments}
\label{sec:experiments}

\subsection{Validating Shift Loss: Artistic Style Transfer for Videos}
\label{sec:experiments-style}

To evaluate our method, we firstly conduct experiments for \textit{artistic style transfer} for videos,
to validate the effectiveness of \textit{shift loss} on constraining consistency for sequential frames.
We collect a training dataset of 98 HD video footage sequences
(from \textit{VIDEVO}\footnote{http://www.videvo.net} containing 2450 frames in total);
the \textit{Sintel} \cite{butler2012naturalistic}
sequences are used for testing, as their ground-truth optical flow is available.
We compare the performance of the models trained under the following setups:
(1) \textit{\textbf{FF}} \cite{johnson2016perceptual}: Canonical feed forward style transfer trained on single frames;
(2) \textit{\textbf{FF+flow}} \cite{huang2017real}: \textit{FF} trained on sequential images, with optical flow added for imposing temporal constraints on subsequent frames;
(3) \textit{\textbf{Ours}}: \textit{FF} trained on single frames, with an additional \textit{shift loss} as discussed in Sec. \ref{sec:shift-loss}.

\begin{figure*}[!ht]
\begin{minipage}{\textwidth}
    \begin{minipage}[b]{0.72\textwidth}
        \centering
        {\includegraphics[width=\textwidth]{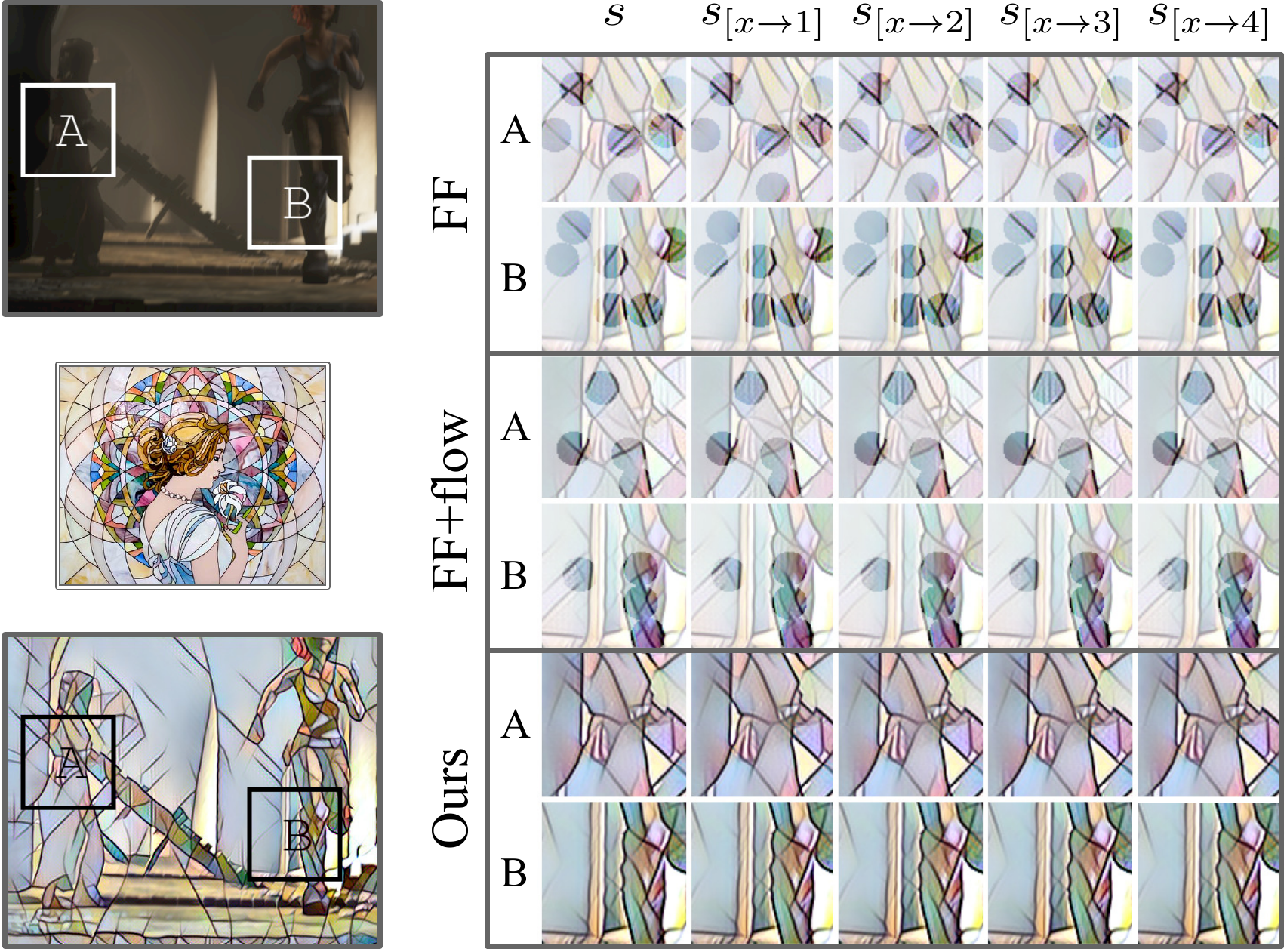}}
        \captionof{figure}{
            Shift-invariance evaluation, comparing between \textit{FF}, \textit{FF+flow} and \textit{Ours}.
            We shift an input image $s$ along the $X$ axis by $1,2,3,4$ pixels respectively and feed all $5$ frames through the networks trained via \textit{FF}, \textit{FF+flow} and \textit{Ours} and show the generated stylizations.
            We mark the most visible differences with small circles and dim the rest of the generated images.
            As is discussed in \protect\cite{ruder2017artistic}, \textit{FF} generates almost identical stylizations for $s$ and $s_{\left[x\rightarrow4\right]}$ (because $4$ is a multiple of the total downsampling factor of the trained network),
            but those for $s_{\left[x\rightarrow1\right]}$,$s_{\left[x\rightarrow2\right]}$,$s_{\left[x\rightarrow3\right]}$ differ significantly.
            \textit{FF+flow} improves the shift-invariance, but we suspect the improvement is due to the inexplicit consistency constraint on regional shifts imposed by optical flow.
            \textit{Ours} is able to generate shift-invariant stylizations with the proposed \textit{shift loss}.
        \label{fig:shift-compare}
        }
    \end{minipage}
    \hfill
    \begin{minipage}[b]{0.26\textwidth}
        \centering
        {\includegraphics[width=\textwidth]{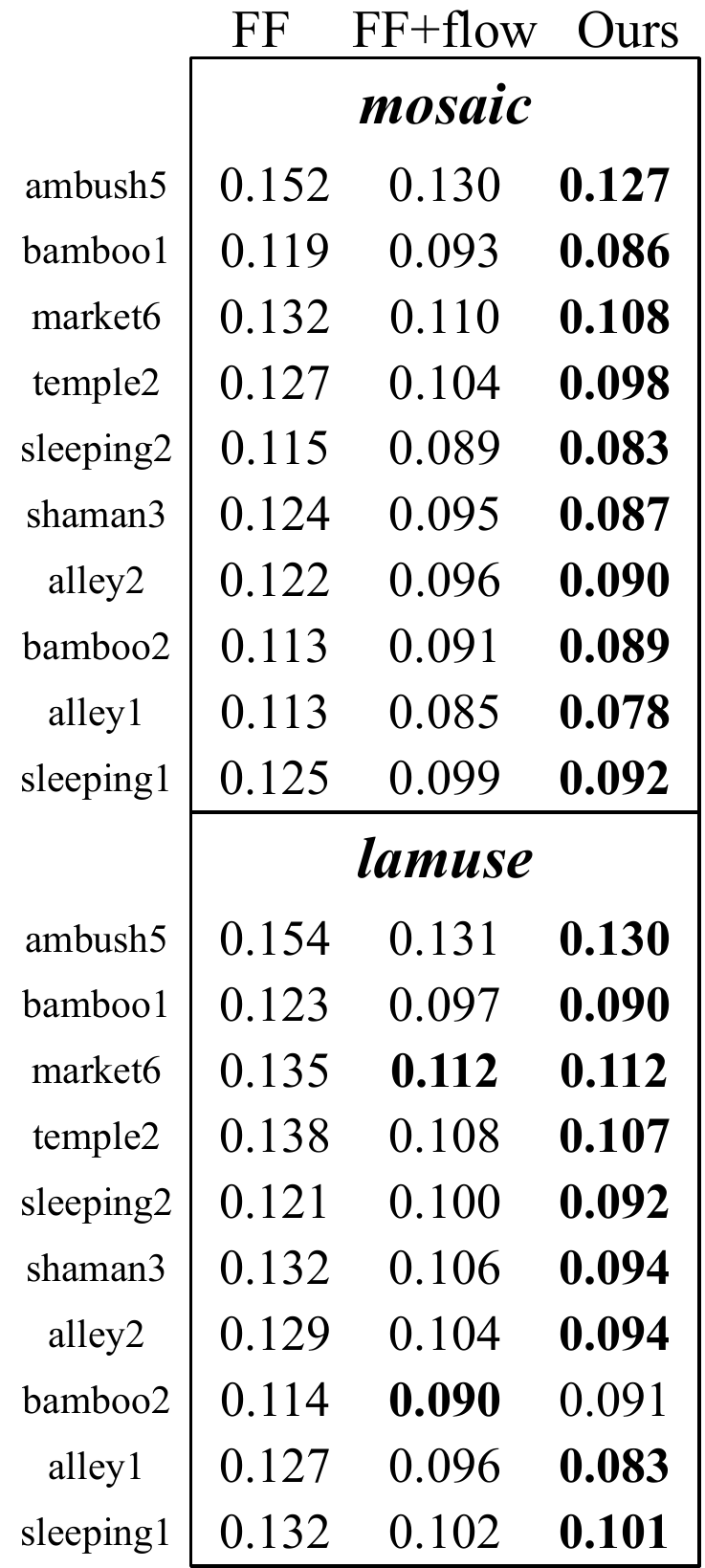}}
        \captionof{table}{
            Comparing temporal loss between \textit{FF}, \textit{FF+flow} and \textit{Ours}.
            \textit{FF+flow} directly optimizes on this metric,
            while optical flow is never provided to \textit{Ours}; yet \textit{Ours} achieves lower temporal loss on the evaluated \textit{Sintel} sequences.
        \label{tab:sintel-flos}
        }
    \end{minipage}
\end{minipage}
\end{figure*}

As a proof of concept, we begin our evaluation by comparing the three setups on their ability to generate shift invariant stylizations for shifted single frames.
In particular, for each image $s$ in the testing dataset, we generate $4$ more test images by shifting the original image along the $X$ axis by $1,2,3,4$ pixels respectively,
and pass all $5$ frames ($s$, $s_{\left[x\rightarrow1\right]}$, $s_{\left[x\rightarrow2\right]}$, $s_{\left[x\rightarrow3\right]}$, $s_{\left[ x\rightarrow 4 \right] }$) through the trained network to examine the consistency of the generated images.
The results shown in Fig. \ref{fig:shift-compare} validate the discussion from \cite{ruder2017artistic}, since the stylizations for $s$ and $s_{\left[x\rightarrow4\right]}$ from \textit{FF} are almost identical ($K=4$ for the trained network), but differ substantially otherwise.
\textit{FF-flow} improves the invariance by a limited amount;
\textit{Ours} is capable of generating consistent stylizations for shifted inputs, with the \textit{shift loss} directly reducing the shift variance.

We then evaluate the consistency of stylized sequential frames,
computing the temporal loss \cite{huang2017real} using the ground truth optical flow for the \textit{Sintel} sequences (Table \ref{tab:sintel-flos}).
Although the temporal loss is part of the optimization objective of \textit{FF-flow},
and our method does not have access to any optical flow information, \textit{Ours} is still able to achieve lower temporal loss with the \textit{shift loss} constraint.

\begin{figure}[!h]
    \centering
        \includegraphics[width=\columnwidth]{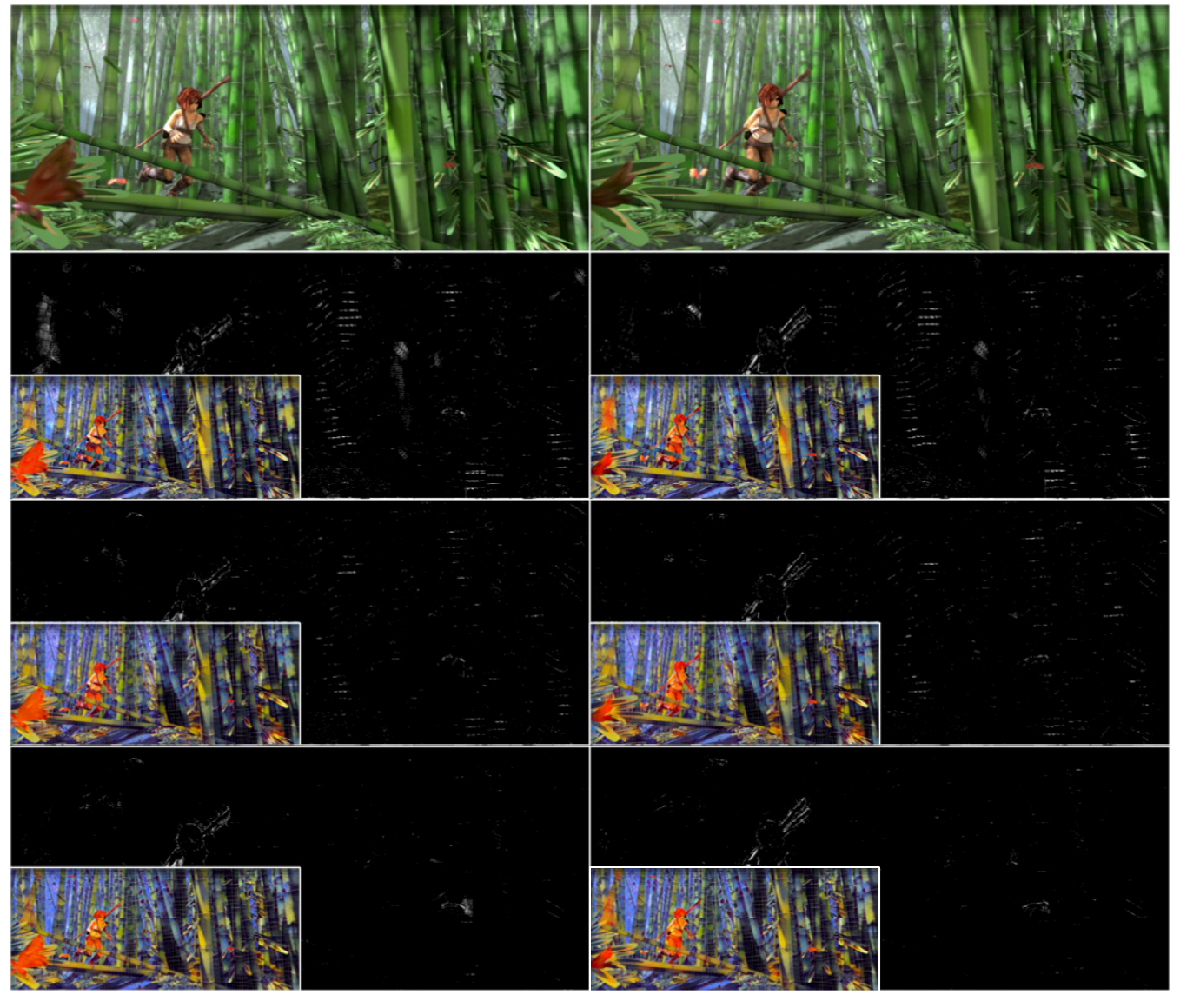}
    \caption{
    \textit{Temporal error maps} between generated stylizations for subsequent input frames.
    The error increases linearly as shown from black to white in grayscale.
    $\nth{1}$ \textit{row}: input frames; $\nth{2}\sim\nth{4}$ \textit{row}: temporal error maps (with the corresponding stylizations shown on top) of outputs from \textit{FF}, \textit{FF+flow}, and \textit{Ours}.
    We here choose a very challenging style (\textit{mosaic}) for temporal consistency, as it contains many fine details, with tiny tiles laid over the original image in the final stylizations.
    Yet, \textit{Ours} achieves very high consistency.}
   \label{fig:shift-sequence}
\end{figure}

We further visualize the consistency comparison in Fig. \ref{fig:shift-sequence}, where
we show the \textit{temporal error maps}, the same metric as in \cite{huang2017real}, of two stylized consecutive frames for each method. The error increases linearly as shown from black to white in grayscale.
\textit{Ours} (bottom row) achieves the highest temporal consistency.
Further details about style transfer training and the calculation of \textit{temporal error map} are available in the supplement file \cite{zhang2018vrsup}.

\subsection{Quantitative Evaluation: Carla Benchmark}
\label{sec:experiments-ral-carla}

\begin{figure}[t]
    \centering
    \begin{subfigure}{0.49\columnwidth}
      \includegraphics[width=\columnwidth]{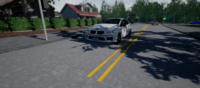}
      \caption{\textit{daytime}}
      \label{fig:carla-a}
    \end{subfigure}
    \begin{subfigure}{0.49\columnwidth}
      \includegraphics[width=\columnwidth]{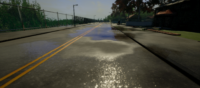}
      \caption{\textit{daytime after rain}}
      \label{fig:carla-b}
    \end{subfigure}
    \begin{subfigure}{0.49\columnwidth}
      \includegraphics[width=\columnwidth]{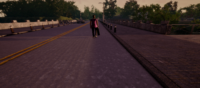}
      \caption{\textit{clear sunset}}
      \label{fig:carla-c}
    \end{subfigure}
    \begin{subfigure}{0.49\columnwidth}
      \includegraphics[width=\columnwidth]{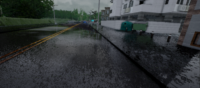}
      \caption{\textit{daytime hard rain}}
      \label{fig:carla-d}
    \end{subfigure}
    \caption{\textit{Carla} weather conditions used in benchmarking: three training conditions (a), (b), (c) and one testing condition (d).}
    \label{fig:carla-weathers}
\end{figure}

Secondly, we conduct a quantitative evaluation of our proposed \textit{real-to-sim} policy transfer pipeline.
Since there are no publicly available common benchmarks for real-world autonomous driving evaluation, we test our pipeline in the \textit{Carla} simulator following its benchmark setup \cite{dosovitskiy2017carla,Codevilla2018}. We choose the imitation learning pipeline because the reinforcement learning policy in \cite{dosovitskiy2017carla} performs substantially worse.
In \cite{dosovitskiy2017carla}, the expert datasets for \textit{Carla} benchmark are collected under $4$ different weather conditions (\textit{daytime}, \textit{daytime after rain}, \textit{daytime hard rain} and \textit{clear sunset}), and the policy is tested on benchmark tasks under \textit{cloudy daytime} and \textit{soft rain at sunset}.
Since the datasets under the testing benchmark conditions are not available\footnote{https://github.com/carla-simulator/imitation-learning} for us to conduct domain adaptation, we split the provided training datasets into three training conditions (\textit{daytime}, \textit{daytime after rain}, \textit{clear sunset}) and one testing condition (\textit{daytime hard rain}) as shown in Fig. \ref{fig:carla-weathers}.

We present comparisons for both phases in the policy transfer pipeline: \textit{policy training} and \textit{domain adaptation}.

For the \textit{policy training} phase, we adopt the following training regimes:
(1) \textit{\textbf{Single-Domain}}: We train one policy under each of the three training weather conditions;
(2) \textit{\textbf{Multi-Domain}}: A policy is trained under a combined dataset containing all three training weather conditions.
We note that since the imitation policy is trained with datasets instead of interacting with the simulation environment, the full approach of \textit{Domain Randomization} \cite{Sadeghi2017cadrl, tobin2017domain} could not be directly applied, as it requires to randomize the textures of each object, lighting conditions and viewing angles of the rendered scenes.
Thus the \textit{Multi-Domain} can be considered as a relatively limited realization of the \textit{Domain Randomization} approach in the \textit{Carla} benchmark dataset setup.
As for the progressive nets approach \cite{rusu2016sim}, it requires a finetuning phase of the policy in the real world, which for autonomous driving means that we need to deploy the trained policy onto a real car and finetune it through rather expensive real-world interactions. Thus we do not consider this approach in this evaluation. (An additional comparison experiment with the progressive nets can be found in the supplementary materials \cite{zhang2018vrsup}.)

\begin{table*}[bh]
\centering
\begin{tabular}{llcccccccc}
\hline
                                                                                                                                          &                  & \multicolumn{2}{c}{Training}                                                                                                                         & \multicolumn{6}{c}{Testing}                                                                     \\ \cline{3-10}
                                                                                                                                          &                  & \multirow{2}{*}{\begin{tabular}[c]{@{}c@{}}Single-\\ Domain\end{tabular}} & \multirow{2}{*}{\begin{tabular}[c]{@{}c@{}}Multi-\\ Domain\end{tabular}} & \multicolumn{3}{c}{Single-Domain}          & \multicolumn{3}{c}{Multi-Domain}                   \\ \cline{5-10}
                                                                                                                                          &                  &                                                                           &                                                                          & No-Gog. & CycleGAN      & VR-Gog.    & No-Gog.         & CycleGAN      & VR-Gog.    \\ \hline
\multirow{4}{*}{\begin{tabular}[c]{@{}l@{}}Success rate \\ (\%)\end{tabular}}                                                             & Straight         & 81.3                                                                      & 97.3                                                                     & 13.3       & \textbf{93.3} & 90.7          & 64.0               & 96.0          & \textbf{100}  \\
                                                                                                                                          & One turn         & 64.0                                                                      & 85.3                                                                     & 1.3        & \textbf{54.7} & \textbf{54.7} & 36.0               & 61.3          & \textbf{76.0} \\
                                                                                                                                          & Navigation       & 60.0                                                                      & 84.0                                                                     & 0.0        & 21.3          & \textbf{45.3} & 0.0                & 42.7          & \textbf{61.3} \\
                                                                                                                                          & Nav. dynamic     & 58.7                                                                      & 74.7                                                                     & 0.0        & 21.3          & \textbf{32.0} & 0.0                & 34.7          & \textbf{56.0} \\ \hline
\multirow{4}{*}{\begin{tabular}[c]{@{}l@{}}Ave. distance \\ to goal \\ travelled \\ (\%)\end{tabular}}                                    & Straight         & 89.7                                                                      & 96.5                                                                     & 37.8       & \textbf{95.8} & 94.7          & 83.6               & 95.9          & \textbf{98.3} \\
                                                                                                                                          & One turn         & 73.6                                                                      & 71.1                                                                     & 18.4       & 36.3          & \textbf{48.4} & 24.7               & 52.0          & \textbf{71.3} \\
                                                                                                                                          & Navigation       & 68.6                                                                      & 88.8                                                                     & 7.3        & 36.7          & \textbf{51.0} & 7.4                & 60.8          & \textbf{73.1} \\
                                                                                                                                          & Nav. dynamic     & 68.2                                                                      & 80.7                                                                     & 5.0        & 32.6          & \textbf{51.3} & 5.2                & 55.7          & \textbf{72.2} \\ \hline
\multirow{5}{*}{\begin{tabular}[c]{@{}l@{}}Ave. distance \\ travelled between\\ two infractions\\ in Nav. dynamic \\ (km)\end{tabular}} & Opposite lane    & 2.83                                                                      & 2.55                                                                     & 0.23       & \textbf{0.77} & 0.72          & 0.26               & 0.83          & \textbf{2.22} \\
                                                                                                                                          & Sidewalk         & 6.47                                                                      & 9.70                                                                     & 0.21       & 1.15          & \textbf{2.62} & 0.38               & 1.29          & \textbf{2.46} \\
                                                                                                                                          & Collision-static & 2.38                                                                      & 3.03                                                                     & 0.14       & 0.52          & \textbf{0.87} & 0.16               & 0.77          & \textbf{1.26} \\
                                                                                                                                          & Collision-car    & 2.06                                                                      & 1.03                                                                     & 0.29       & 1.01          & \textbf{1.40} & 0.27               & 0.59          & \textbf{0.77} \\
                                                                                                                                          & Collision-pedestrian   & 15.10                                                                     & 16.17                                                                    & 2.17       & 4.03          & \textbf{6.98} & \textgreater{}4.60 & \textbf{7.29} & 4.67          \\ \hline
\end{tabular}
\caption{Quantitative evaluation of goal-directed \textit{Carla} navigation benchmark tasks \cite{dosovitskiy2017carla}. We train imitation policies under single weather condition (\textit{Single-Domain}) and three training weather conditions (\textit{Multi-Domain}). Policies are evaluated in testing weather condition through direct deploying (\textit{No-Goggles}), translating the input image through \textit{CycleGAN} and through \textit{VR-Goggles} (for transferring the \textit{Multi-Domain} policy with \textit{CycleGAN} and \textit{VR-Goggles} we train one adaptation network from the testing weather condition to each of the three training weather conditions and report the average results under those three adaptations). Higher is better.}
\label{tab:carla_metric}
\end{table*}

For the \textit{domain adaptation} phase, we compare the following adaptation methods:
(1) \textit{\textbf{No-Goggles}}: Feed the testing data directly to the trained policy;
(2) \textit{\textbf{CycleGAN}} \cite{zhu2017unpaired}: Use \textit{CycleGAN} to translate the test data to the training domain before feeding
to policy nets and
(3) \textit{\textbf{Ours}}:
Add \textit{shift loss} on top of (2) as \textit{\textbf{VR-Goggles}} to translate the inputs.
For both \textit{CycleGAN} and \textit{VR-Goggles}, we train an adaptation network from the testing weather condition to each of the three training conditions.
(For more details about the training of the policy and adaptation models, please refer to the supplementary materials \cite{zhang2018vrsup}.)

The four benchmark tasks (\textit{Straight}, \textit{One Turn}, \textit{Navigation} and \textit{Nav. dynamic}) are in order of increasing difficulty and each of them consists of 25 different preset trajectories.
Since the \textit{Multi-Domain} policy is trained with three weather conditions instead of four as in the original setup \cite{dosovitskiy2017carla} due to the reason discussed earlier, directly deploying the \textit{Multi-Domain} policy fail to finish any of the two harder tasks under the relatively extreme testing weather condition.
For the different adaptation strategies, our \textit{VR-Goggles} outperforms \textit{CycleGAN} on almost all of the metrics, especially the two harder tasks (\textit{Navigation} and \textit{Nav. dynamic}) in terms of both the success rate and the average percentage of distance to goal traveled.
The average distance traveled between two infractions is reported only for the hardest task \cite{dosovitskiy2017carla}: navigating in the presence of dynamic objects (\textit{Nav. dynamic}).
The adaptation models of \textit{Ours} enable the agents to drive safely with mostly lower infraction frequencies compared with \textit{CycleGAN}.
\textit{CycleGAN} collides with pedestrians less often with the \textit{Multi-Domain} policy.
A probable explanation is that most episodes under this setup end due to collision with cars and static obstacles, so there does not occur too many challenging pedestrian conditions.
For example, for direct deployment without adaptation (\textit{No-Goggles}), the average distance between collisions with pedestrians is higher than 4.6 km, because the total navigation distance for all 25 episodes in this task is only 4.6 km which is too short to encounter pedestrians.

We note that the transfer pipelines of \textit{Single-Domain} policies behave much better than directly deploying the \textit{Multi-Domain} policy, and the training time of the former policy is also much shorter than that of the latter \cite{zhang2018vrsup}.

\subsection{Real-world Indoor \& Outdoor Navigation}
\label{sec:ral-real}

\begin{figure*}[!th]
    \includegraphics[width=\textwidth]{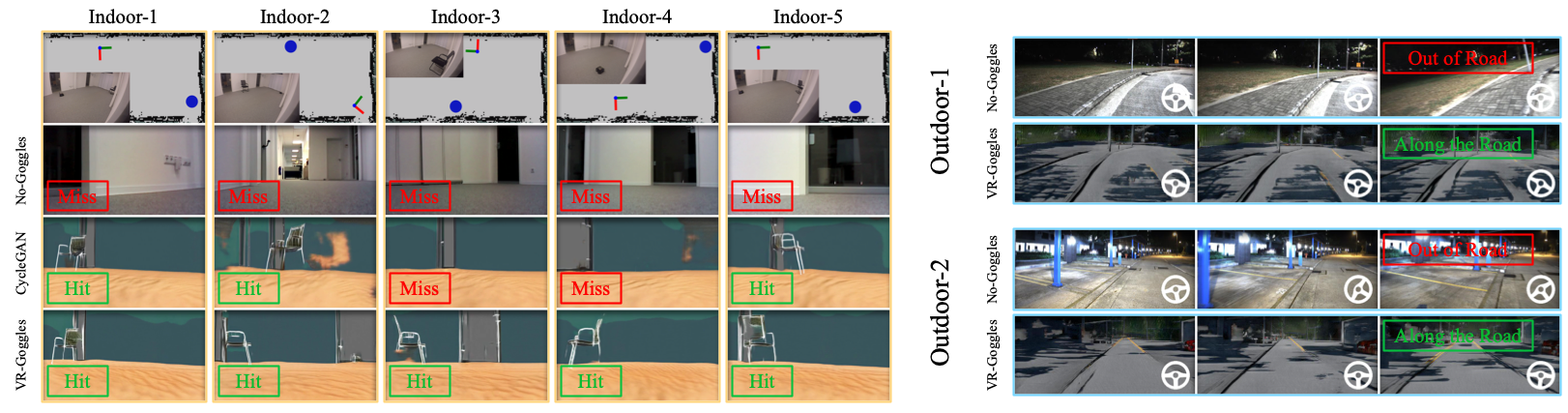}
    \label{fig:realworld}
  \caption{
Real-world visual control experiments.
\textit{\textbf{Indoor}} (yellow):.
A navigation policy is firstly trained in a simulated environment
(Fig. \ref{fig:in-sim})
that is able to navigate to chairs based on visual inputs.
Without retraining or finetuning,
our proposed \textit{VR-Goggles} enables the mobile robot to
directly deploy this policy in a real office environment (Fig. \ref{fig:in-real}),
achieving $100\%$ success rate in a set of real-world experiments.
Here \textit{Miss} refers to test runs
where the agent stays put or rotate in place and simply ignores the chair even when they are in sight as the policy trained in the simulation could not cope with the drastically visually different inputs (\textit{No-Goggles}),
or due to the inconsistency of the translated subsequent outputs which hinders the successful fulfilment of the goal-reaching task (\textit{CycleGAN}).
\textit{Hit} refers to frames where the agent captures the chair in sight
and outputs commands to move towards it.
\textit{\textbf{Outdoor}} (cyan):
An autonomous driving policy (via conditional imitation learning \cite{Codevilla2018})
is trained in \textit{Carla} daytime (Fig. \ref{fig:out-sim}),
a \textit{VR-Goggles} model is trained to translate
between \textit{Carla} daytime and \textit{Robotcar} nighttime (Fig. \ref{fig:out-real}),
which enables the real-world nighttime deployment of the trained policy.
  }
  \label{fig:realworld}
\end{figure*}

\begin{figure}[!t]
    \centering
    \begin{subfigure}{0.49\columnwidth}
    \centering
        \includegraphics[width=\columnwidth]{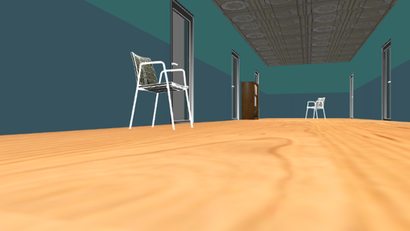}
        \caption{simulated indoor}
        \label{fig:in-sim}
    \end{subfigure}
    \begin{subfigure}{0.49\columnwidth}
    \centering
        \includegraphics[width=\columnwidth]{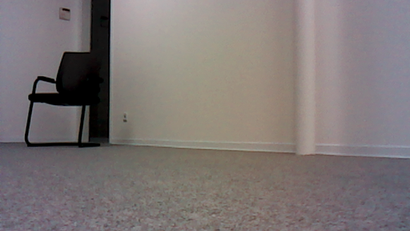}
        \caption{real indoor}
        \label{fig:in-real}
    \end{subfigure}
    \begin{subfigure}{0.49\columnwidth}
    \centering
        \includegraphics[width=\columnwidth]{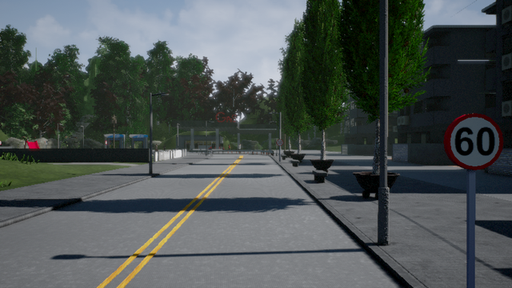}
        \caption{simulated outdoor}
        \label{fig:out-sim}
    \end{subfigure}
    \begin{subfigure}{0.49\columnwidth}
    \centering
        \includegraphics[width=\columnwidth]{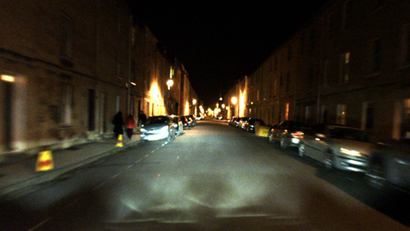}
        \caption{real outdoor}
        \label{fig:out-real}
    \end{subfigure}
    \caption{Samples from the simulated environment (left) and the real world (right) used in our indoor (top) and outdoor (bottom) navigation experiments.}
    \label{fig:sim-real}
\end{figure}

Finally, we conduct real-world robotics experiments for both indoor and outdoor visual navigation tasks.
We begin by training learning-based visual navigation policies,
taking simulated first-person-view images as inputs,
outputting moving commands for specific navigation targets.
Then, we deploy the trained policy onto real robots,
comparing the following \textit{domain adaptation} approaches:
(1) \textit{\textbf{No-Goggles}}: Feed the sensor readings directly to the trained policy;
(2) \textit{\textbf{CycleGAN}}/\textit{\textbf{CyCADA}} \cite{zhu2017unpaired,hoffman2017cycada}: Use \textit{CycleGAN} (when semantic ground truth is not available) / \textit{CyCADA} (when ground truth semantic maps are provided by the simulator) to translate the real sensory inputs to the synthetic domain before feeding
to the policy nets;
(3) \textit{\textbf{Ours}}:
Add \textit{shift loss} on top of (2)
as the \textit{\textbf{VR-Goggles}}.

For \textit{\textbf{indoor}} office experiments,
we build an office environment in \textit{Gazebo} \cite{koenig2004design}
and render $s \sim p_{\text{sim}}$ from this simulation environment (Fig. \ref{fig:in-sim}).
We capture $r \sim p_{\text{real}}$ from a real office (Fig. \ref{fig:in-real})
using a \textit{RealSense R200 camera} mounted on a \textit{Turtlebot3 Waffle}.
For conducting the \textit{domain adaptation},
as the simulator (\textit{Gazebo}) does not provide ground truth semantics,
we drop the semantic constraint $\mathcal{L}_{\text{sem}}$.
The input images are of size $640\times360$
and the adaptation network is trained with $256\times256$ crops.
We use the same network architecture as in \textit{CycleGAN},
and train for 50 epochs with a learning rate of $2e-4$
as we observe no performance gain training for longer iterations.

We train the navigation policy using Canonical A3C with 8 parallel workers \cite{mnih2016asynchronous} in \textit{Gazebo},
and deploy the trained policy
onto \textit{Turtlebot3 Waffle}
and compare the three \textit{domain adaptation} approaches
(Fig. \ref{fig:realworld}).
Without \textit{domain adaptation},
\textit{No-Goggles}
fails completely in the real-world tasks;
our proposed \textit{VR-Goggles} achieves the highest success rate
($0\%$, $60\%$ and $100\%$ for \textit{No-Goggles},
\textit{CycleGAN} and \textit{Ours} respectively) due to the quality and consistency of the translated streams.
The control cycle runs in real-time at $13\Hz$ on a \textit{Nvidia TX2}.

Finally, we conduct \textit{\textbf{outdoor}} autonomous driving experiments
(we sample $s \sim p_{\text{sim}}$ from the \textit{Carla} daytime \cite{dosovitskiy2017carla} environment Fig. \ref{fig:out-sim}
and sample $r \sim p_{\text{real}}$ from a nighttime dataset of \textit{Robotcar} \cite{RobotCarDatasetIJRR} Fig. \ref{fig:out-real})
with input images of size $640\times400$.
Considering that \textit{VR-Goggles} outperforms \textit{CycleGAN} in
\textit{indoor} experiments,
and since outdoor robotics experiments are relateively expensive,
we only compare \textit{No-Goggles} and \textit{VR-Goggles} in the \textit{outdoor} autonomous driving scenario.
We take the driving policy trained through conditional imitation learning \cite{Codevilla2018} as in Section \ref{sec:experiments-ral-carla}.
This policy takes as inputs the first person view RGB image and a high-level command,
which falls in a discrete action space and is generated through a global planner (\textit{straight, left, right, follow, none}).
In our real-world experiments,
this high-level direction command is set as \textit{straight},
indicating the vehicle
(a \textit{Bulldog} with a \textit{PointGrey Blackfly} camera mounted on it)
to always go along the road.
The control policy outputs the steering angle.

The control policy is trained purely in \textit{Carla} simulated daytime,
while it is tested in a nighttime town street scene (Fig. \ref{fig:realworld}).
It is non-trivial to quantitatively evaluate the control policy in the real world,
so we show two representative sequences marked with the output steering commands.
The top row of each sequence shows the continuous outputs of \textit{No-Goggles}.
Due to the huge difference between the real nighttime and the simulated daytime,
the vehicle failed to move along the road.
Our \textit{VR-Goggles}, however,
successfully guides the vehicle along the road as instructed by the global planner
(the policy prefers to turn right since it is trained in a right-driving environment)
\footnote{A video demonstrating our approach and much more experimental results are available at \url{https://sites.google.com/view/zhang-tai-19ral-vrg/home}, where we also show that the \textit{VR-Goggles} can easily train a new model for a new type of chair without
finetuning the indoor control policy.}.



\section{Conclusions}
\label{sec:conclusion}

In this paper, we tackle the \textit{reality gap} occurring when deploying learning-based visual control policies trained in simulation to the real world, by translating the
real images
back to the synthetic domain during deployment. Due to the sequential nature of the incoming sensor streams for control tasks, we propose \textit{shift loss} to increase the consistency of the translated subsequent frames, and validate it both in \textit{artistic style transfer for videos} and \textit{domain adaptation}.
We verify our proposed \textit{VR-Goggles} pipeline as a lightweight, flexible and efficient solution for visual control through \textit{Carla} benchmark as well as a set of real-world robotics experiments.
It would be interesting to apply our method to manipulation,
as this paper has been mainly focused on navigation.
Also, evaluating our method in more challenging environments
on more sophisticated control tasks could be another future direction.


\section*{Acknowledgment}
The authors would like to thank Christian Dornhege and Daniel Büscher for the discussion of the initial idea.


\bibliographystyle{IEEEtran}
\bibliography{tai19icra}


\end{document}